\pdfoutput=1

\documentclass[11pt]{article}

\usepackage[final]{acl}

\usepackage{times}
\usepackage{latexsym}

\usepackage[T1]{fontenc}

\usepackage[utf8]{inputenc}

\usepackage{microtype}

\usepackage{inconsolata}

\usepackage{graphicx}

\usepackage{algorithm}
\usepackage[utf8]{inputenc} 
\usepackage[T1]{fontenc}    
\usepackage{url}            
\usepackage{booktabs}       
\usepackage{amsfonts}       
\usepackage{nicefrac}       
\usepackage{microtype}      
\usepackage{bm}
\usepackage{enumitem}
\usepackage{graphicx}
\usepackage{amsmath}
\usepackage{subfigure}
\usepackage{amssymb}
\usepackage{caption}

\usepackage{comment}
\usepackage{array}
\usepackage{multirow}
\usepackage{wrapfig,lipsum,booktabs}
\usepackage{algpseudocode}
\usepackage{makecell}
\usepackage{pifont}
\usepackage{tabularx}
\usepackage{color,colortbl}
\definecolor{Gray}{gray}{0.9}

\usepackage{longtable}

\newcommand{\skrnospace}{\texttt{Ski}}
\newcommand{\skr}{\texttt{Ski} }
\newcommand{\skrqn}{\texttt{Ski-Q-n} }
\newcommand{\skrqnnospace}{\texttt{Ski-Q-n}}
\newcommand{\skrqcn}{\texttt{Ski-QC-n} }
\newcommand{\skrqcnnospace}{\texttt{Ski-QC-n}}
\newcommand{\skrqan}{\texttt{Ski-QA-n} }
\newcommand{\skrqannospace}{\texttt{Ski-QA-n}}

\newcommand{\skrqcannospace}{\texttt{Ski-QCA-n}}

\newcommand{\skrqcassemble}{\texttt{Ski-QC-ASM} }
\newcommand{\skrqcassemblenospace}{\texttt{Ski-QC-ASM}}
\newcommand{\skrqaassemble}{\texttt{Ski-QA-ASM} }
\newcommand{\skrqaassemblenospace}{\texttt{Ski-QA-ASM}}
\newcommand{\skrqcaassemble}{\texttt{Ski-QCA-ASM} }
\newcommand{\skrqcaassemblenospace}{\texttt{Ski-QCA-ASM}}

\newcommand{\skrqone}{\texttt{Ski-Q-1} }

\newcommand{\skrqonenospace}{\texttt{Ski-Q-1}}
\newcommand{\skrqcone}{\texttt{Ski-QC-1} }
\newcommand{\skrqctwo}{\texttt{Ski-QC-2} }
\newcommand{\skrqcthree}{\texttt{Ski-QC-3} }
\newcommand{\skrqconenospace}{\texttt{Ski-QC-1}}
\newcommand{\skrqaone}{\texttt{Ski-QA-1} }

\newcommand{\skrqaonenospace}{\texttt{Ski-QA-1}}
\newcommand{\skrqcaone}{\texttt{Ski-QCA-1} }
\newcommand{\skrqcaonenospace}{\texttt{Ski-QCA-1}}
\newcommand{\skrqcatwo}{\texttt{Ski-QCA-2} }
\newcommand{\skrqcathree}{\texttt{Ski-QCA-3} }

\newcommand{\skrcone}{\texttt{Ski-C-1} }
\newcommand{\skrconenospace}{\texttt{Ski-C-1}}
\newcommand{\skrcassemble}{\texttt{Ski-C-ASM} }
\newcommand{\skrcassemblenospace}{\texttt{Ski-C-ASM}}

\title{Synthetic Knowledge Ingestion: Towards Knowledge Refinement and Injection for Enhancing Large Language Models}

\author{Jiaxin Zhang$^{1,2}$, \ Wendi Cui$^2$, \ Yiran Huang$^2$, \ Kamalika Das$^{1,2}$, \ Sricharan Kumar$^{1,2}$ \\ $^1$Intuit AI Research \quad $^2$Intuit  \\
\texttt{\{jiaxin\_zhang, wendi\_cui, yiran\_huang, kamalika\_das, sricharan\_kumar\}@intuit.com}}

\begin{document}
\maketitle
\begin{abstract}

Large language models (LLMs) are proficient in capturing factual knowledge across various domains. However, refining their capabilities on previously seen knowledge or integrating new knowledge from external sources remains a significant challenge. In this work, we propose a novel synthetic knowledge ingestion method called \skrnospace, which leverages fine-grained synthesis, interleaved generation, and assemble augmentation strategies to construct high-quality data representations from raw knowledge sources. We then integrate \skr and its variations with three knowledge injection techniques: Retrieval Augmented Generation (RAG), Supervised Fine-tuning (SFT), and Continual Pre-training (CPT) to inject and refine knowledge in language models. Extensive empirical experiments are conducted on various question-answering tasks spanning finance, biomedicine, and open-generation domains to demonstrate that \skr significantly outperforms baseline methods by facilitating effective knowledge injection. We believe that our work is an important step towards enhancing the factual accuracy of LLM outputs by refining knowledge representation and injection capabilities.
\footnote{The source code and datasets will be publicly available for research purposes.}

\end{abstract}
\section{Introduction}

Large language models (LLMs) demonstrate proficiency in capturing vast amounts of factual information across a wide range of fields, attributable to their extensive pre-training datasets \cite{petroni2019language, cohen2023crawling, hu2023survey}. Although these models hold an impressive repository of knowledge, integrating new information via external datasets or enhancing their capacity on previously seen information still poses several challenges. One primary challenge is {\em outdated knowledge}, where the static nature of the information fails to evolve over time. Another issue is the {\em domain knowledge deficit} where language models, typically generalists, lack detailed, specialized knowledge in sectors like finance \cite{wu2023bloomberggpt} and healthcare \cite{singhal2023large}. Additionally, there is the problem of {\em catastrophic forgetting}, where language models may lose previously acquired knowledge \cite{luo2023empirical}, which 
particularly affects rare facts that are minimally represented in the training data \cite{kandpal2023large}. These issues underscore the necessity for ongoing enhancements to the knowledge capabilities.

\begin{figure}[t]
    \includegraphics[width=0.99\linewidth]{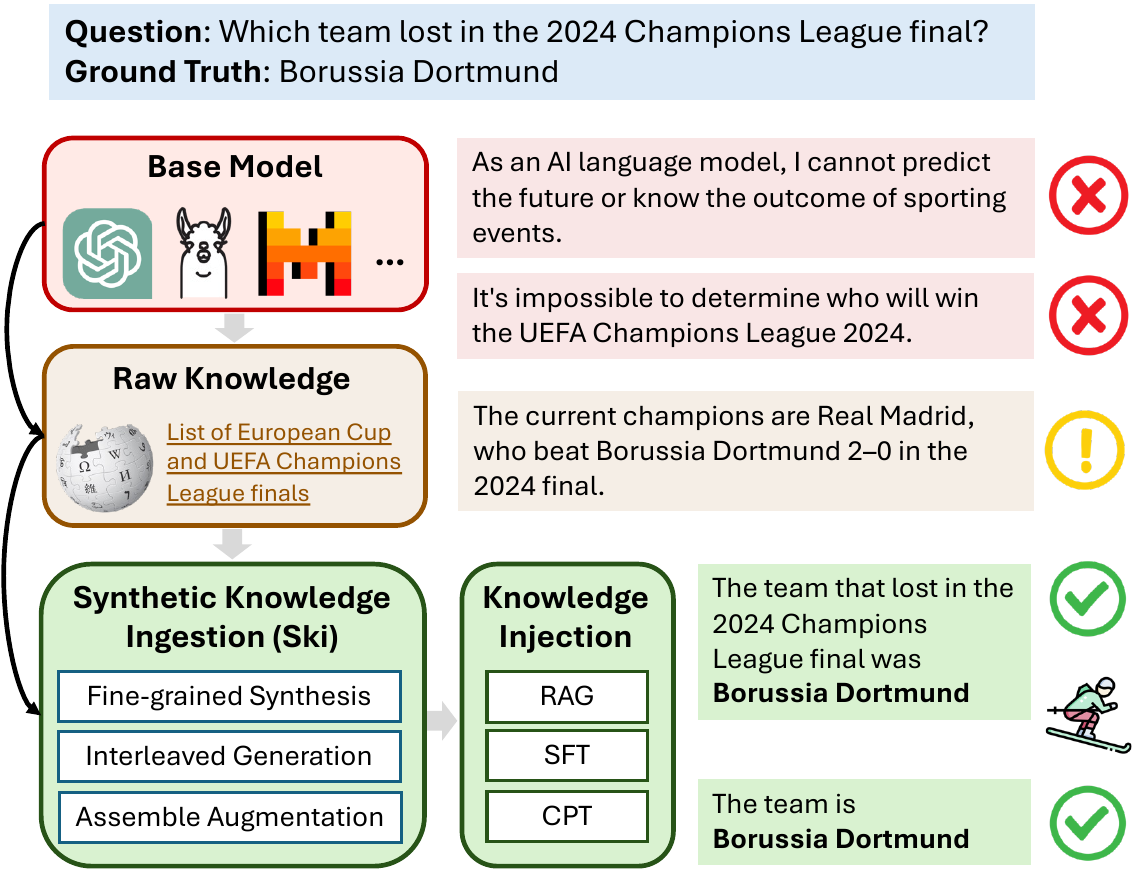}
    \caption{Illustrative example. Base models are often unable to handle certain questions due to limited knowledge. Even with raw knowledge provided, the output answer may still be incorrect if not well-digested by LLMs. Our proposed Synthetic Knowledge Ingestion method, \skrnospace, incorporates three key innovations to easily transform raw knowledge into refined data representations that LLM can effectively digest. By utilizing injection pipelines such as RAG, SFT, and CPT, knowledge or information will be injected into LLM to ensure accurate and correct answers.}
    \label{fig:illustrative_example}
\end{figure}

\begin{figure*}[ht]
\centering
    \includegraphics[width=0.99\textwidth]{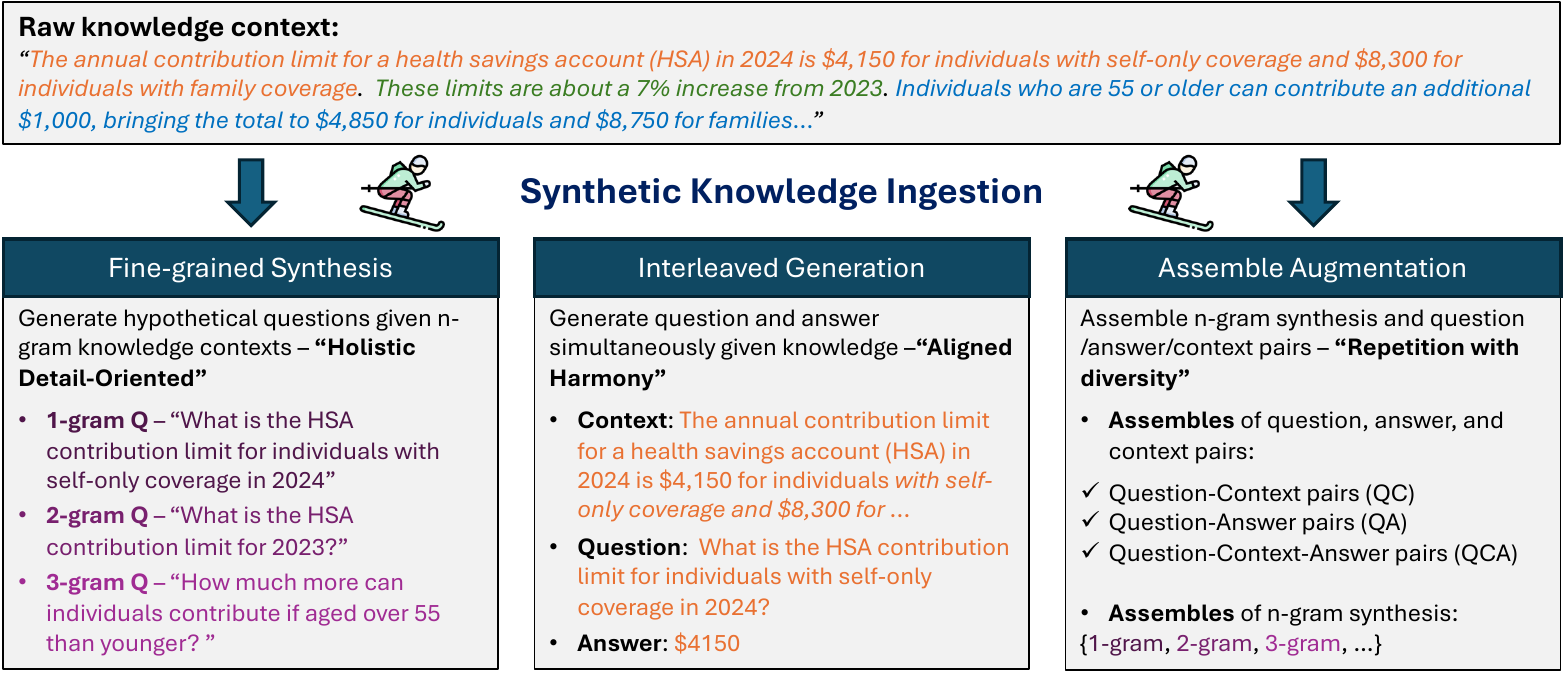}
    \caption{Overview of the proposed method: Synthetic Knowledge Ingestion (\skrnospace), comprises of three essential components. Firstly, fine-grained synthesis incorporates generating hypothetical questions based on an $n$-gram detail-oriented principal. Secondly, interleaved generation generates questions and answers simultaneously by maintaining aligned harmony, given a specific knowledge piece. Lastly, assemble augmentation combines question, answer, and context pairs, along with $n$-gram synthesis, to improve the repetition with diversity. }
    \label{fig:overview_method}
\end{figure*}


In response to emerging challenges, the practice of customizing Large Language Models (LLMs) to specialized fields, coupled with regular updates to their knowledge base, is becoming increasingly prevalent \cite{yu2022survey,wang2023knowledge}. Various strategies have been developed to enhance the factual accuracy and domain-specific expertise of LLMs. These initiatives are crucial for integrating precise and current knowledge, thus broadening the utility and effectiveness of LLMs within professional settings. Despite these advances, the performance of LLMs often remains suboptimal, with issues such as hallucination \cite{manakul2023selfcheckgpt, zhang2023sac3}, especially in tasks that require extensive knowledge. Consequently, assessing the LLM's capability to understand, memorize, and retrieve knowledge is substantially essential. Given some knowledge in the form of a raw text corpus, our objective is to understand two key concepts and fundamental questions: 
(i) {\bf Knowledge injection}: What is the optimal approach to encoding targeted knowledge into a language model to enhance its capability and functionality? and (ii) {\bf Knowledge ingestion}: What are the most effective strategies for formatting knowledge or databases to ensure that such representations are readily assimilable and digestible by large language models (LLMs) prior to injection?\footnote{Ingestion centers on the acquisition and processing of knowledge for storage and future use, whereas injection focuses on actively integrating knowledge to enhance a language model, either during inference or training phases.} 


A prevalent technique for injecting external information into model responses is Retrieval Augmented Generation ({\bf RAG}) \cite{lewis2020retrieval}. This method relies on external knowledge sources, yet its effectiveness is contingent upon the efficacy of its retrieval system \cite{chen2024benchmarking}. If this retrieval process is flawed, it might not secure the necessary information, resulting in inappropriate responses. Additionally, a misalignment in knowledge representation could hinder RAG's comprehension of the context \cite{gao2023retrieval}. Another approach is Supervised Fine-Tuning ({\bf SFT}) \cite{dettmers2024qlora}, which improves a model by continuing its training on relevant tasks. Instruction tuning \cite{zhang2023instruction,taori2023alpaca} under SFT has shown substantial enhancements in language model performance, though it doesn't necessarily enrich the model's knowledge base \cite{ouyang2022training,chia2023instructeval,zhou2024lima}. Techniques in fine-tuning also include reinforcement learning (RL) strategies such as RLHF \cite{touvron2023llama,achiam2023gpt} and DPO \cite{rafailov2024direct}, which refine the model’s alignment post-training but do not expand its knowledge capacity. Hence, developing a precise supervised mechanism for raw knowledge implementation poses a significant challenge. A third strategy is Continual Pre-Training ({\bf CPT}) \cite{ke2023continual}, or unsupervised fine-tuning, where the model is further trained on specific knowledge datasets tailored to certain tasks or domains \cite{wu2024continual}. While CPT does not require labeled data, structuring the data in a format that closely reflects the specific goals and tasks can be particularly effective.
Recent studies have garnered attention for these three knowledge injection strategies \cite{ovadia2023fine,balaguer2024rag,mecklenburg2024injecting}, yet there remains a gap in knowledge ingestion. Given unprocessed and unstructured knowledge, two key questions arise: (i) {\em What data representations are most effective for each injection strategy?} (ii) {\em How can we systematically construct diverse and high-quality representations to facilitate effective knowledge injection?}


Inspired by this gap, we introduce \underline{S}ynthetic \underline{k}nowledge \underline{i}ngestion (\skrnospace), an innovative synthetic data generation method that automates and enhances knowledge ingestion, as illustrated in Fig.\ref{fig:illustrative_example}. \skr leverages three key innovations to generate high-quality and diverse data representation from a raw knowledge base. First, {\em Fine-grained Synthesis} creates hypothetical questions based on $n$-gram knowledge contexts, ensuring a detailed match in relevance, minimizing the semantic gap between questions and answers, and increasing representation diversity. Second, {\em Interleaved Generation} simultaneously generates both questions and answers based on specific knowledge. This synthetic question-answering (QA) format naturally mirrors the process of information-seeking, providing direct contextual alignment and relevance between the questions and their respective answers. Third, {\em Assemble Augmentation} combines fine-grained synthesis across different $n$-gram spans and their QA pair iterations, balancing repetition with diverse elements. By integrating these components, the \skr approach significantly enhances the refinement of knowledge from its raw state, thus facilitating effective knowledge ingestion into LLMs.

\skr offers a generic solution for generating synthetic data from raw knowledge, which can be integrated with all three knowledge injection strategies: RAG, SFT, and CPT.
With this advancement, \skr not only incorporates new information using external datasets but also refines LLMs' capability on previously seen knowledge and information.  We provide a comprehensive evaluation of \skr on two open-source LLMs, \texttt{Llama2-7B} \cite{touvron2023llama} and \texttt{Mistral-7B} \cite{jiang2023mistral}, over cross-domain QA tasks, including those in the finance, biomedical, open-generation and multi-hop domain. Through performing extensive experiments, our approach \skr achieves substantial improvements upon the baselines by a large margin. In summary, our contributions are threefold:
\begin{itemize}[leftmargin=10pt,nosep]
    \vspace{5pt}
    \item We develop \skrnospace, a novel synthetic knowledge ingestion method that integrates three essential innovations, advancing the capability of refining knowledge representation from raw contexts.
    \vspace{5pt}
    \item We integrate \skr with three knowledge injection strategies—RAG, SFT, and CPT—to effectively inject and refine knowledge in LLMs. 
    \vspace{5pt}
    \item We conduct experiments on various question-answering tasks across different domains. Our results demonstrate that \skr significantly outperforms baseline methods by a large margin.
\end{itemize}

\section{Problem Formulation}

\paragraph{Knowledge Ingestion.} It generally refers to the methodology of acquiring, integrating, and transforming information from a variety of knowledge sources. It encompasses gathering, absorbing, and morphing knowledge to build a database that facilitates future use. Mathematically, let's $\mathcal{M}$ represent a language model, $\mathcal{Q}$ denote a set of factual questions, and assume that we only have access to relevant raw knowledge base $\mathcal{K}_\mathcal{Q}$. We aim to enhance the representation of the knowledge base, represented as  $\mathcal{K}^*_{\mathcal{Q}}$ 
\begin{equation}
    \mathcal{K}^*_{\mathcal{Q}} = \mathcal{T}(\mathcal{K}_\mathcal{Q}), \quad \mathcal{T}(\cdot): \textup{text} \rightarrow \textup{text}
\end{equation}
where $\mathcal{T}$ signifies a transformation process that converts raw knowledge into refined knowledge $\mathcal{K}_\mathcal{Q} \rightarrow \mathcal{K}^*_{\mathcal{Q}}$. Although advanced knowledge representations such as knowledge graphs show promise, their discussion falls beyond the scope of this study. 

\paragraph{Knowledge Injection.} It involves actively encoding or integrating specific knowledge into a pre-trained language model to enhance its performance by incorporating new information from external datasets or refining the model's capabilities on previously seen information. Specifically, when the refined knowledge applied improves the understanding of the pre-trained model $\mathcal{M}$ regarding questions $\mathcal{Q}$ through a transformation $\mathcal{F}$,
\begin{equation}
    \mathcal{M}^*:=\mathcal{F}(\mathcal{M}, \mathcal{K}^*_{\mathcal{Q}}) \quad \textup{s.t.} \quad \mathcal{S}_{\mathcal{M}^*, {\mathcal{Q}}} > \mathcal{S}_{\mathcal{M}, \mathcal{Q}}
\end{equation}

where $\mathcal{S}$ is a score metric, e.g., an accuracy that measures the performance of the pre-trained model in answering the questions $\mathcal{Q}$, and $\mathcal{M}^*$ is the updated language model. We explore three options for $\mathcal{F}$: RAG, SFT, and CPT, and provide a systematic evaluation of how our proposed synthetic knowledge ingestion method, \skrnospace, can enhance the capabilities of LLMs across various QA tasks.



\section{Synthetic Knowledge Ingestion}
The human learning process often involves asking questions and inquiring ``why'', which can enhance comprehension. Drawing inspiration from this, \skr also leverages the power of questions. By utilizing three key innovations, \skr transforms raw knowledge (such as documents and articles) into question-augmented representations. The following section provides a detailed description of these three key innovations, as illustrated in Figure \ref{fig:overview_method}.
 
\subsection{Fine-grained Synthesis} 
Considering the scenario where we possess a paragraph of text as a knowledge base, crafting effective questions to transform and augment this knowledge base poses a challenging task. Questions that are too broad may struggle to encompass all pertinent knowledge points, whereas excessively detailed questions risk losing sight of the overall content. Moreover, ensuring that these questions are both non-repetitive and diverse adds extra challenges. 

To tackle these challenges, we introduce a fine-grained synthesis approach, inspired by n-gram language models, allowing for a balanced incorporation of both detailed and overarching content. Assume the knowledge base $\mathcal{K}_{\mathcal{Q}}$ paragraph consists of $m$ sentences, $\mathcal{K}_{\mathcal{Q}} = \{k_1, k_2, ..., k_m \}$ where $k_i$ means the $i$-th sentence in $\mathcal{K}_{\mathcal{Q}}$, we generate hypothetical questions by querying an LLM model, given a specific set of sentences conditioning on the whole knowledge paragraph $\mathcal{K}_{\mathcal{Q}}$:
\begin{equation}
   q_j^n \leftarrow \mathcal{L}_{P_{fs}}(\{k_j, k_{j+1},...,k_{j+n-1}\}; \mathcal{K}_{\mathcal{Q}}),
\end{equation}
where $\mathcal{L}$ is a pioneer LLM model with meta-prompt $P_{fs}$,  $q_j^n$ is the $j$-th $(1 \le j \le m-n+1)$ generated hypothetical questions given the specific sentence set $\{k_j, k_{j+1},...,k_{j+n-1}\}$ and the raw knowledge context $\mathcal{K}_{\mathcal{Q}}$. We denote $\Tilde{Q}^n$ as the set of $n$-gram hypothetical questions:
    $\Tilde{Q}^n = \{q_1^n, q_2^n, q_3^n,...,q_{m-n}^n\}$.
We define a set of $n$-gram knowledge context $C^n$ 
\begin{equation}
\begin{aligned}
    C^n = & \{ \{k_1,...,k_n\},..., \{k_{m-n+1}, k_m \} \}
\end{aligned}
\end{equation}
and therefore have $C^1 = \{k_1,k_2,...,k_m\}$, $C^2 = \{ \{k_1, k_2\}, \{k_2, k_3\},...,\{k_{m-1}, k_m \} \}$, etc. The question-context pair $\{\Tilde{Q}^n, C^n \}$ can be written by:
\begin{equation}
\begin{aligned}
    \{\Tilde{Q}^n, C^n \} = &\{ \{q_1^n, \{ k_1,...,k_n\} \},...,
    \\&\{q_{m-n}^n,\{k_{m-n+1}, k_m \} \} \}.
\end{aligned}
\end{equation}
Specifically, we have two variants of \skr leveraging $n$-gram synthesis principle: 
\begin{itemize}[leftmargin=10pt]
    \item \skrqnnospace: synthetic hypothetical questions $\Tilde{Q}^n$ by $n$-gram given knowledge contexts $C^n$
    \item \skrqcnnospace: synthetic hypothetical questions with knowledge context pair $\{\Tilde{Q}^n, C^n \}$ by $n$-gram
\end{itemize}
where $n$ typically ranges from 1 to 3. Specifically, $\skrqone$ refers synthetic 1-gram questions over $m$ sentences, $\Tilde{Q}^1 = \{q_1^{1}, q_2^{1},...,q_m^{1}\}$, and $\skrqconenospace$ means synthetic 1-gram question-context (QC) pair $ \{\Tilde{Q}^1, C^1 \} = \{\{q_1^{1}, k_1\},..., \{q_m^{1},k_m\} \}$ that includes $m$ pairs in this set. 
More details and examples can be found in the Appendix \ref{sec:ngram}. 





\subsection{Interleaved Generation} 





Hypothetical question and context pairs have the potential to transform knowledge articles into questions (\skrqnnospace) and question-context (QC) pairs (\skrqcnnospace), which can revolutionize representations of knowledge articles. However, SFT requires succinct QA pairs, while the context portion of the QC pair consists of sentences, which are too lengthy to be formulated as answers.

To deal with this issue, we introduce an interleaved generation strategy that simultaneously generates QA pairs based on a specific knowledge context. This synthetic QA format naturally emulates the information-seeking process, providing direct contextual alignment and relevance between the questions and their corresponding answers. This strategy can be built upon hypothetical questions yet delivers QA pairs tailored to the specific knowledge context $\mathcal{K}_{\mathcal{Q}}$:
\begin{equation}
   \{q_j^n, a_j^n \} \leftarrow \mathcal{L}_{P_{ig}}(\{k_j,...,k_{j+n-1}\}; \mathcal{K}_{\mathcal{Q}}),
\end{equation}
where $P_{ig}$ is the meta-prompt (see details in Appendix) different from the prompt $P_{fs}$ used above. We denote $\Tilde{A}^n$ as the set of answers corresponding to the $n$-gram hypothetical questions:
    $\Tilde{A}^n = \{a_1^n, a_2^n, a_3^n,...,a_{m-n}^n\}$. 
The question-answering pairs $\{\Tilde{Q}^n, A^n \}$ can be written by:
\begin{equation}
    \{\Tilde{Q}^n, \Tilde{A}^n \} = \{ \{q_1^n, a_1^n\},...,\{q_{m-n}^n,a_{m-n}^n  \} \}.
\end{equation}

The synthetic questions $\Tilde{Q}^n$ and corresponding answers $\Tilde{A}^n$ align well with the given knowledge context $C^n$ due to interleaved mechanism. We thus denote the following \skr variants from the perspective of the QA pair:
\begin{itemize}[leftmargin=10pt]
    \item \skrqannospace: synthetic QA pair $\{\Tilde{Q}^n, \Tilde{A}^n\}$ by $n$-gram given specific knowledge contexts $C^n$
    \item \skrqcannospace: synthetic QA pair combined with the knowledge context, $\{\Tilde{Q}^n, C^n, \Tilde{A}^n\}$
\end{itemize}

\subsection{Assemble Augmentation} 

To enhance the effectiveness of knowledge ingestion, we propose a strategic assembly approach that harnesses the benefits of fine-grained and interleaved generation. This strategy is built upon two main pillars: (1) article/document augmentation for optimized retrieval, and (2) data augmentation for both supervised and unsupervised fine-tuning. Specifically, we first combine all question-context pairs from the same document into a singular article, thereby improving the retrieval quality
\begin{equation}
\begin{aligned}
    [\Tilde{Q}C]^n = &\{q_1^n, \{ k_1,...,k_n\} \} \oplus \{q_2^n, \{ k_2,...,k_{n+1}\} \} \\ &...\oplus \{q_{m-n}^n,\{k_{m-n+1}, k_m \} \} 
\end{aligned}
\end{equation}
where $\oplus$ denotes the concatenation operator, which combines all QC pair strings into a single article. The notation $[\Tilde{Q}C]^n$ represents the augmented article composed of all QC pairs using $n$-gram techniques. 
As the same information can exist in different formats under different $n$-gram strategies, combining pairs from different $n$-gram allows repetition and diversity, enhancing the depth and breadth of knowledge integration. 

Assemble augmentation also offers scalability suitable for synthesizing expansive sets of question-answer pairs utilizing various $n$-gram generation techniques.
For fine-tuning schemes, this strategy can be employed to generate a diverse ensemble of $n$-gram QA pairs for data augmentation purposes. Specifically, the QA assembly aggregates all $n$-gram QA pairs as follows:
\begin{equation}
    [\Tilde{{Q}} \Tilde{{A}}]^{n} = 
    \{\Tilde{Q}^1, \Tilde{A}^1\} \cup...\cup\{\Tilde{Q}^n, \Tilde{A}^n\}
\end{equation}
where $[\Tilde{{Q}} \Tilde{{A}}]^{n}$ is a combination of QA pairs from $1$-gram to $n$-gram. Similarly, we have the ensemble of QCA pairs from 1-gram to $n$-gram:
\begin{equation}
    [\Tilde{{Q}}C\Tilde{{A}}]^{n} = 
    \{\Tilde{Q}^1, C^1, \Tilde{A}^1\} \cup...\cup\{\Tilde{Q}^n, C^n, \Tilde{A}^n\},
\end{equation}
where $\cup$ denotes the union of sets, which refers to containing all the elements in a large set. We thus summarize three \skr variants from the perspective of assemble that typically performs article augmentation (for retrieval) or data augmentation:  
\begin{itemize}[leftmargin=10pt]
    \item \skrqcassemblenospace: assemble of all question-context pairs into  one augmented set $ [\Tilde{Q}C]^n$ 
    \item \skrqaassemblenospace: assemble of all question-answer pairs from $1$-gram to $n$-gram $[\Tilde{{Q}} \Tilde{{A}}]^{n}$
    \item \skrqcaassemblenospace: assemble of all question-context-answer pairs from $1$-gram to $n$-gram $[\Tilde{{Q}}C\Tilde{{A}}]^{n}$
\end{itemize}

\begin{table*}[!h]
\resizebox{\linewidth}{!}{
\footnotesize
\begin{tabular}{@{}c|l|cccc|cccc|cccc@{}}
\toprule
\multirow{2}{*}{\bf Model}      & \multirow{2}{*}{\bf Method} & \multicolumn{4}{c|}{\bf BioASQ}          & \multicolumn{4}{c|}{\bf NQ}              & \multicolumn{4}{c}{\bf FiQA}            \\ \cmidrule(l){3-14} 
                            &                         & nDCG@1 & nDCG@10 & Recall@1 & Recall@10 & nDCG@1 & nDCG@10 & Recall@1 & Recall@10 & nDCG@1 & nDCG@10 & Recall@1 & Recall@10 \\ \midrule
\multirow{3}{*}{Contriever} & Raw Article                   & 0.810   & 0.720   & 0.133   & 0.578     & 0.748  & 0.831  & 0.644   & 0.915     & 0.298   & 0.323   & 0.134   & 0.406       \\
                            & \skrqone (iHyDE)                & \bf{0.845}  & 0.733  & \bf{0.142}   & 0.574     & 0.752  & 0.804  & 0.641   & 0.876     & 0.452   & 0.482   & 0.231   & 0.562       \\ 
                            \rowcolor{Gray}%
                            \cellcolor{white}%
                            & \skrqcone              & 0.835  & \bf{0.751}  & 0.132   & \bf{0.597 }    & \bf{0.810}   & \bf{0.865}  & \bf{0.704}   & \bf{0.926}     & \bf{0.455}   & \bf{0.493}   & \bf{0.232}   & \bf{0.584}        \\
                            \rowcolor{Gray}
                            \cellcolor{white}%
                            & \skrqcassemble       & 0.820   & 0.719  & 0.131   & 0.575     & 0.773  & 0.839  & 0.670    & 0.908     & 0.398   & 0.435   & 0.193   & 0.542        \\ 
                            \midrule
\multirow{3}{*}{BM25}       & Raw Article                    & 0.232  & 0.297  & 0.189   & 0.389     & 0.325  & 0.389  & 0.272   & 0.478     & 0.315   & 0.318   & 0.162   & 0.368     \\
                            & \skrqone (iHyDE)                & 0.175  & 0.219  & 0.142   & 0.289     & 0.218  & 0.282  & 0.176   & 0.368     & 0.238   & 0.252   & 0.121   & 0.307      \\ 
                            \rowcolor{Gray}%
                            \cellcolor{white}%
                            & \skrqcone               & 0.210   & 0.282  & 0.173   & 0.373     & 0.3    & 0.377  & 0.245   & 0.482     & 0.289   & 0.301   & 0.142   & 0.362       \\
                            \rowcolor{Gray}%
                            \cellcolor{white}%
                            & \skrqcassemble      & \bf{0.268}  & \bf{0.328}  & \bf{0.221}   & \bf{0.417}     & \bf{0.350}   & \bf{0.422}  & \bf{0.291}   & \bf{0.513}     & \bf{0.377}   & \bf{0.390}    & \bf{0.192}   & \bf{0.454}        \\ 
                            \bottomrule
\end{tabular}
}
\vspace{-2mm}
\caption{Retrieval performance on three datasets using two different retrievers (Contriever and BM25). }
\label{tab:retrieval}
\end{table*}


\begin{figure*}[h!]
\centering
    \includegraphics[width=0.32\textwidth]{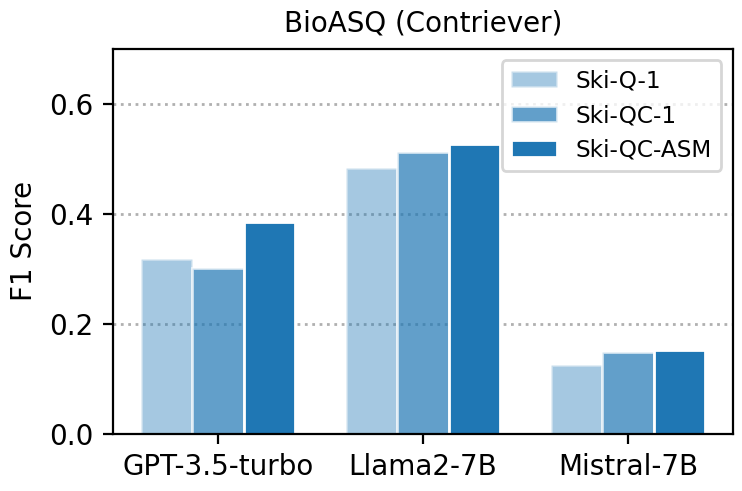}
    \includegraphics[width=0.32\textwidth]{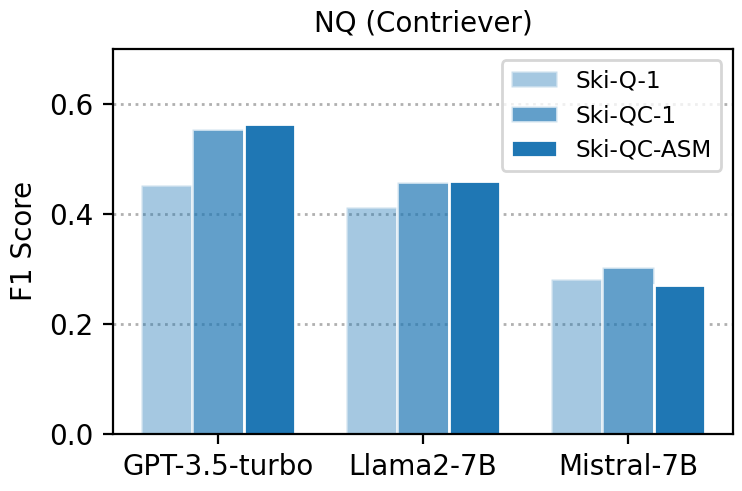}
    \includegraphics[width=0.32\textwidth]{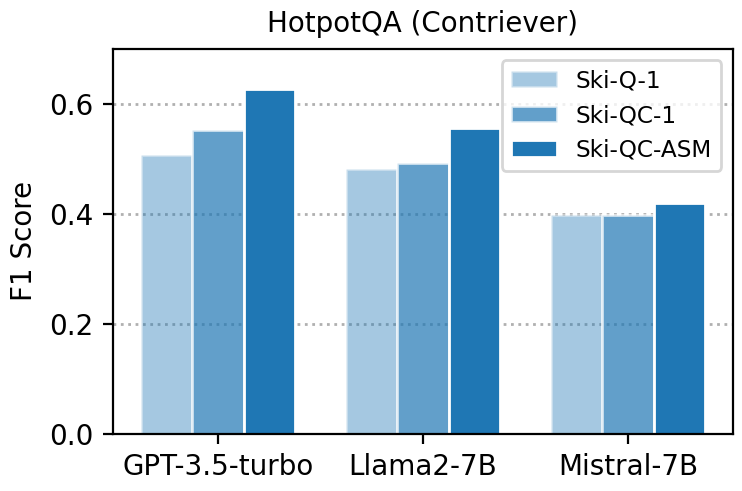}
    \includegraphics[width=0.32\textwidth]{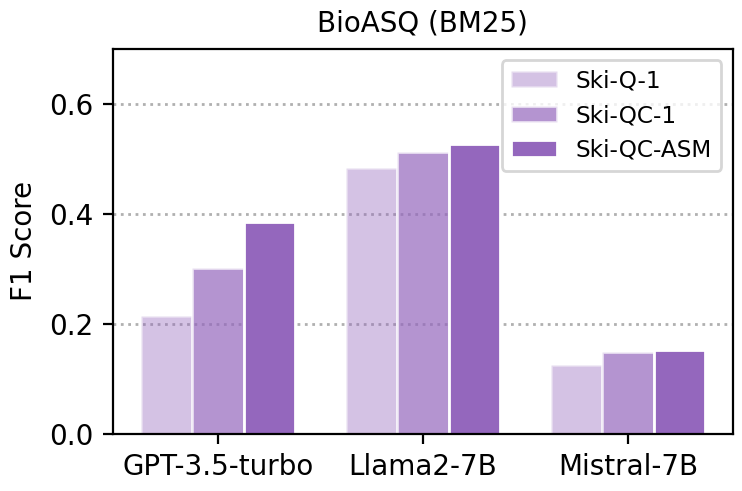}
    \includegraphics[width=0.32\textwidth]{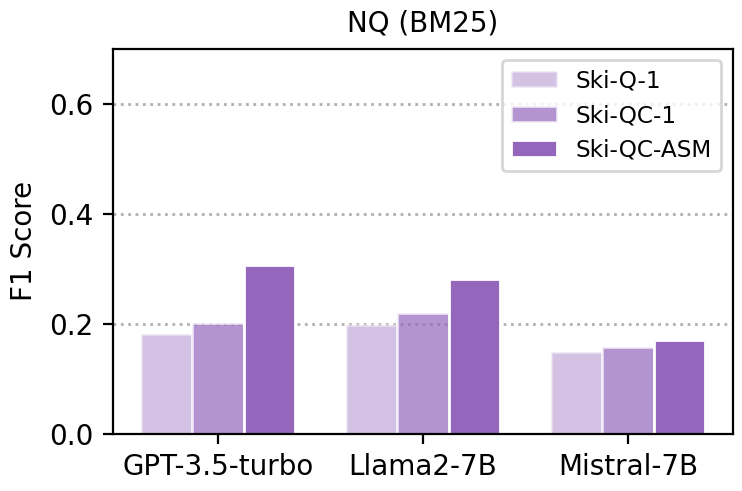}
    \includegraphics[width=0.32\textwidth]{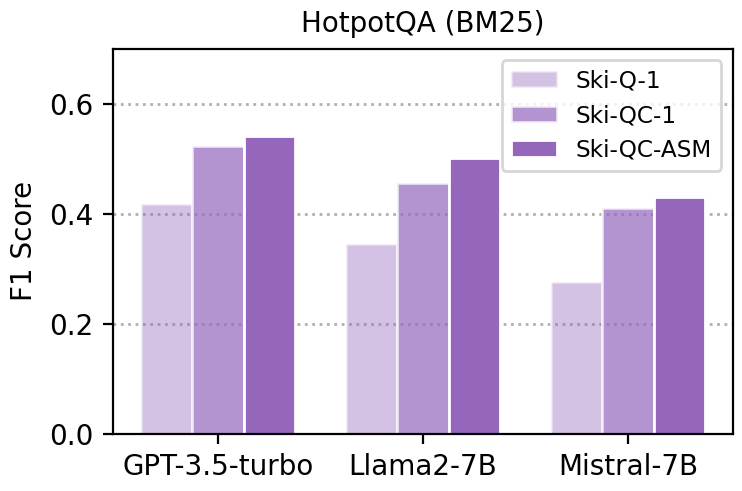}
    \vspace{-2mm}
    \caption{Effect of \skr method on the RAG performance using different retrievers and generators. }
    \label{fig:rag_effect}
\end{figure*}

\section{Knowledge Injection for Enhancing Language Models}

\subsection{Retrieval Augmented Generation (RAG)}

We employ three variations of the \skr approach, namely \skrqnnospace, \skrqcnnospace, and \skrqcassemblenospace, when utilizing the RAG pipeline for knowledge injection. At the retrieval stage, \skrqn matches the query's embedding with embeddings of all questions generated in an $n$-gram pattern from the knowledge documents.
This can be interpreted as an inverse HyDE implementation, where HyDE \cite{gao2022hyde} transfers a query to a potential answer and performs an answer-to-knowledge search. For \skrqn we transform knowledge into questions and execute a question-to-question (Q-Q) search. In contrast, \skrqcn compares the query's embedding with embeddings of all question-context (QC) pairs, facilitating a question-to-(question+context) (Q-QC) search. Additionally, \skrqcn introduces another variation, \skrqcassemble, where QC pairs from the same document are amalgamated into a single article, and the query's embedding is compared with embeddings of such articles. After identifying the most relevant items, \skr consolidates only the most pertinent snippets or sentences from the knowledge documents related to the retrieved items into an LLM for generating the final answer. This process minimizes noise from sentences in the same documents but not directly pertinent to user queries.


\subsection{Supervised Fine-tuning (SFT)}
We utilize three variants of the \skr method, namely question-answer (QA) pairs, question-context (QC) pairs, and question-context-answer (QCA) pairs, to incorporate knowledge via supervised fine-tuning. In the QCA approach, we concatenate the question (Q) and context (C) input and use the answer as the output, following the setting in \citet{viswanathan2023prompt2model}. To conform to the $n$-gram principle, we create separate training datasets for every $n$-gram ($n$=1,2,3) for SFT, resulting in a decrease in dataset size as $n$ increases. We focus on evaluating three fine-grained generations ($n$=1), i.e., \skrqaonenospace, \skrqconenospace, and \skrqcaonenospace. 
\subsection{Continual Pre-training (CPT)}

Continual pre-training can leverage the dataset used for SFT, but it typically requires a larger volume of data to enhance the capabilities of LLMs \cite{lin2024rho,ke2023continual}. To address this, we amalgamate all the generated pairs from SFT to create a comprehensive augmented training set, \skrqcassemblenospace, \skrqaassemblenospace, \skrqcaassemblenospace. This strategy not only amplifies repetition but also preserves diversity \cite{ovadia2023fine}. 


\section{Experiments}

\subsection{Experiment Setup}
\paragraph{Datasets.} We evaluate our approach on multiple cross-domain question-answering tasks, including those in the finance domain (e.g., FiQA \cite{maia201818}), biomedical domain (e.g., BioASQ \cite{tsatsaronis2015overview}), open-generation domain (e.g., NQ \cite{kwiatkowski2019natural} and multi-hop domain (e.g., HotpotQA \cite{yang2018hotpotqa}). Please find more relevant details about data usage for RAG, SFT, and CPT in the Appendix \ref{sec:sft}-\ref{sec:cpt}. 
\paragraph{Evaluation Models and Metrics.} 
We utilize \texttt{GPT-3.5-turbo} from OpenAI as our target agent model to generate synthetic questions, answers, and their corresponding pairs. To evaluate our approach, we employ two open-source LLMs, \texttt{Llama2-7B} \cite{touvron2023llama} and \texttt{Mistral-7B} \cite{jiang2023mistral}, which serve as the base models for RAG, SFT and CPT, implemented using the LLaMA-Factory pipeline\footnote{\url{https://github.com/hiyouga/LLaMA-Factory}}. Our evaluation encompasses a comparison between two retrieval methods: the Dense-Retrieval approach, Contriever \cite{izacard2021unsupervised}, and the Sparse-Retrieval method, BM25. We assess the performance of three generators, \texttt{GPT-3.5-turbo}, \texttt{Llama2-7B}, and \texttt{Mistral-7B} using the F1 score as a measure of accuracy, following the RAGGED \cite{hsia2024ragged} pipeline\footnote{\url{https://github.com/neulab/ragged}} across RAG, SFT, and CPT scenarios.
\paragraph{Baseline Methods.} 

For retrieval and RAG tasks, we utilize raw knowledge — specifically, the original article as a standard baseline. Our approach draws inspiration from Hypothetical Document Embeddings, known as HyDE \cite{gao2022precise}. We introduce an adaptation termed "inverse" HyDE (iHyDE) \cite{gao2023retrieval} established by \skrqonenospace. In SFT evaluations, the base model without fine-tuning serves as the standard baseline. We also employ a vanilla QA baseline, which generates a set of question-answer pairs by querying LLMs using extracted documents/articles, as outlined in \cite{mecklenburg2024injecting, balaguer2024rag}. For the CPT pipeline, our standard baselines encompass the base model and raw context, which is aligned with the training dataset setup as detailed in \cite{ovadia2023fine}. Additionally, we integrate two novel baseline methods: \skrconenospace, concentrating uniquely on the 1-gram knowledge context, and \skrcassemblenospace, an ensemble approach that integrates knowledge contexts ranging from 1-gram to 3-gram. Both are variants of the \skr approach.


\begin{figure*}[t]
\centering
    \includegraphics[width=0.32\textwidth]{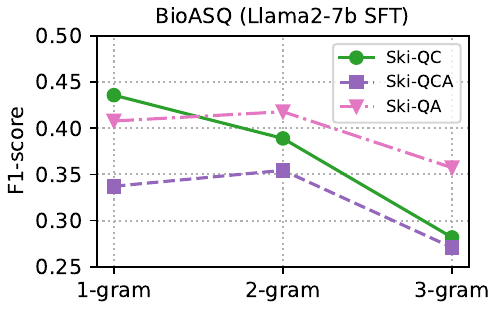}
    \includegraphics[width=0.32\textwidth]{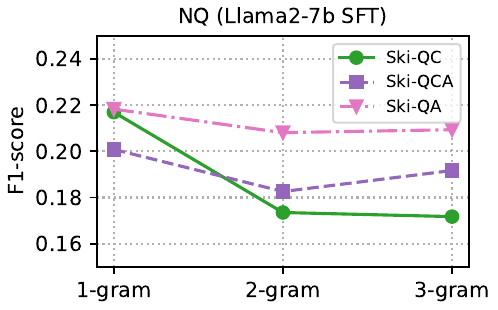}
    \includegraphics[width=0.32\textwidth]{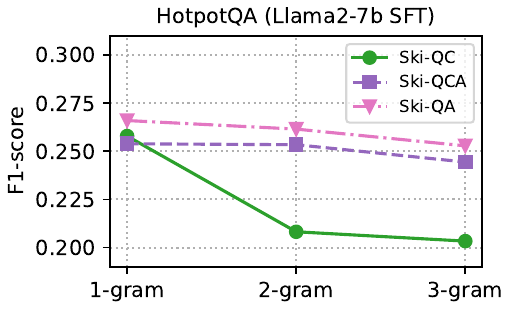}
    \vspace{-2mm}
    \caption{SFT performance on different $n$-gram settings using \texttt{Llama2-7B} base model across three datasets. Note that the F1-score often decreases with the increase of $n$-gram, but not very significantly drop. }
    \label{fig:sft_ngram}
\end{figure*}

\subsection{Main Results}

\paragraph{RAG Results.} 
We evaluate the retrieval performance of \skr using two different retrievers, as illustrated in Table \ref{tab:retrieval}. For Contriever, \skrqcone outperformed both the inverse HyDE (\skrqonenospace) and the raw article baselines, particularly achieving a significant improvement (+15\%) on the FiQA task. For BM25, \skrqcassemble surpassed all the baselines and \skr variations across three datasets.  Further evaluations were conducted on our method within RAG systems, detailed in Table \ref{tab:rag}. Inverse HyDE fell short of the raw article baseline, indicating that the question-to-question search approach might not be as effective as anticipated. \skrqcassemble emerged as the superior method, showing consistent improvements compared to other baselines across all tasks. The results of BM25 are aligned with the retrieval-only task, confirming the effectiveness of assemble augmentation.   

\begin{table}[!h]
\resizebox{\linewidth}{!}{
\begin{tabular}{@{}c|l|ccc@{}}
\toprule
\multirow{1}{*}{\bf Model}                                                                  & \multirow{1}{*}{\bf Method}     & {\bf BioASQ}  & {\bf NQ}  & {\bf HotpotQA} \\ \midrule
\multirow{2}{*}{\begin{tabular}[c]{@{}c@{}}Contriever  
\\ +GPT-3.5     \end{tabular}} & Raw Article     & 0.361       & 0.504   & 0.585       \\
       & \skrqone (iHyDE)               & 0.317       & 0.452   & 0.507        \\
       \rowcolor{Gray}%
       \cellcolor{white}%
       & \skrqcone              & 0.300       & 0.553   & 0.553        \\
       \rowcolor{Gray}%
       \cellcolor{white}%
       & \skrqcassemble      & \bf{0.385}       & \bf{0.563}   & \bf{0.627}        \\ 
        \midrule
\multirow{2}{*}{\begin{tabular}[c]{@{}c@{}}BM25
\\ +GPT-3.5    \end{tabular}}  & Raw Article   & 0.227       & 0.258   & 0.443        \\ 
       & \skrqone (iHyDE)               & 0.162       & 0.181   & 0.274        \\
       \rowcolor{Gray}%
       \cellcolor{white}%
       & \skrqcone               & 0.186      & 0.201   & 0.377        \\
       \rowcolor{Gray}%
       \cellcolor{white}%
       & \skrqcassemble     & \bf{0.243}       & \bf{0.306}   & \bf{0.478}        \\  
     \bottomrule
\end{tabular}
}
\vspace{-2mm}
\caption{End-to-end RAG results with two retrievers.}
\label{tab:rag}
\end{table}

\paragraph{Effect of Different Retrievers and Generators.}  
Fig.\ref{fig:rag_effect} presents a detailed evaluation of how different retrievers and generators influence our \skr methods. In the NQ and HotpotQA tasks, \skr methods that incorporate Contriever generally outperform those using BM25. In terms of generation, \texttt{GPT-3.5-turbo} demonstrates superior performance over \texttt{Llama2-7B} and \texttt{Mistral-7B}. For the BioASQ task, both Contriever and BM25 perform comparably, with \texttt{Llama2-7B} surpassing the other two generators in effectiveness. We compare three variations of \skrnospace, namely \skrqonenospace, \skrqconenospace, and \skrqcassemble, across different combinations of retrievers and generators. Consistently across tasks, \skrqcassemble tends to perform the best, followed by \skrqconenospace, outperforming the baseline \skrqonenospace.

\paragraph{SFT Results.} 

Beyond the RAG evaluation, we assess the enhancement of pre-trained models using SFT in our study with \skrnospace. The synthetic QA pairs, \skrqaonenospace, exhibit superior performance across three datasets. Specifically, \texttt{Mistral-7B} achieves an average performance gain of 5.13\% over the base model and 4.07\% improvement compared to the vanilla QA baseline, while \texttt{Llama2-7B} markedly improves by 19.3\% and 11.1\% respectively. 
Generally, the data representation provided by \skrqaone is the more effective method. We also observe that \texttt{Llama2-7B} demonstrates a stronger capability in refining knowledge compared to \texttt{Mistral-7B}, aligning with findings reported by \citet{ovadia2023fine}.

\begin{table}[!h]
\resizebox{\linewidth}{!}{
\begin{tabular}{@{}c|l|ccc@{}}
\toprule
\multirow{1}{*}{\bf Model}      & \multirow{1}{*}{\bf Method}  & {\bf BioASQ} & {\bf NQ}    & {\bf HotpotQA} \\ \midrule
\multirow{3}{*}{\texttt{Mistral-7B}} & Base Model               & 0.125  & 0.098 & 0.113    \\
                            & Vanilla QA               & 0.131  & 0.084 & 0.155    \\
                               \rowcolor{Gray}%
                            \cellcolor{white}%
                            & \skrqaone                    & {\bf 0.162}  & {\bf 0.159} & {\bf 0.171}    \\
                       \rowcolor{Gray}%
                            \cellcolor{white}%
                            & \skrqcone                    & 0.138  & 0.113 & 0.155    \\
                           \rowcolor{Gray}%
                            \cellcolor{white}%
                            & \skrqcaone                   & 0.160  & 0.157 & 0.163    \\ \midrule
\multirow{3}{*}{\texttt{Llama2-7B}}  & Base Model               & 0.123  & 0.082 & 0.136    \\
                            & Vanilla QA                   & 0.235  & 0.150 & 0.203    \\
                           \rowcolor{Gray}%
                            \cellcolor{white}%
                            & \skrqaone                    & 0.357  & {\bf 0.218} & {\bf 0.266}    \\
                           \rowcolor{Gray}%
                            \cellcolor{white}%
                            & \skrqcone                    & {\bf 0.436}  & 0.217 & 0.258    \\
                           \rowcolor{Gray}%
                            \cellcolor{white}%
                            & \skrqcaone                   & 0.271  & 0.201 & 0.252    \\ \bottomrule
\end{tabular}
}
\vspace{-2mm}
\caption{SFT performance on two pre-trained models. }
\label{tab:sft}
\end{table}

\paragraph{Effect of $n$-gram Synthesis.} 

The impact of $n$-gram on SFT is illustrated in Fig. \ref{fig:sft_ngram}. Across various tasks, the 1-gram consistently outperforms the 2-gram and 3-gram in all \skr variations, underscoring the significance of fine-grained generation for enhancing both the quantity and quality of data synthesis. Compared to QA pairs and QCA pairs, the performance of QC pairs drops slightly and is inferior to the other two variations. This suggests that the interleaved generation in QA pairs is crucial for improving SFT performance.


\paragraph{CPT Results.} 

\texttt{Llama2-7B} demonstrates a superior capability for knowledge injection compared to \texttt{Mistral-7B} in the SFT evaluation. Building on this observation, we further explore the impact of \skr through the CPT approach with \texttt{Llama2-7B}. While QA pairs excel in SFT, \skrqcassemble and \skrqcaassemble show significantly enhanced performance in the CPT evaluation. Specifically, these methods lead to performance gains of +21.2\% in BioASQ and +11.5\% in NQ task relative to the base model. \skrcassemble presents a slight improvement over other QC and QCA variations in HotpotQA. This highlights the benefit of assembly augmentation where longer contexts, which include more comprehensive information than short answers, aid in deepening the pretrained models' understanding.


\begin{table}[!h]
\centering
\resizebox{\linewidth}{!}{
\begin{tabular}{@{}l|ccc@{}}
\toprule
\multirow{1}{*}{\bf Method}  
                        & {\bf BioASQ} & {\bf NQ}    & {\bf HotpotQA} \\ \midrule
Base Model            & 0.123  & 0.082  & 0.136    \\
Raw Article (only context)         & 0.219  & 0.166 & 0.242    \\
\skrcone (1-gram context)             & 0.208  & 0.178 & 0.204    \\
\skrcassemble (1-3-gram context)              & 0.241  & 0.163 & {\bf 0.253}    \\
   \rowcolor{Gray}%
\skrqaone                    & 0.269  & 0.184 & 0.205    \\
   \rowcolor{Gray}%
\skrqaassemble                  & 0.228  & 0.194 & 0.218    \\
   \rowcolor{Gray}%
\skrqcone                    & 0.294  & 0.178 & 0.226    \\
   \rowcolor{Gray}%
\skrqcassemble                  & {\bf 0.335}  & 0.182 & 0.215    \\   \rowcolor{Gray}%
\skrqcaone                   & 0.211  & 0.188 & 0.218    \\
   \rowcolor{Gray}%
\skrqcaassemble                 & 0.220   & {\bf 0.197} & 0.191    \\ \bottomrule
\end{tabular}
}
\vspace{-2mm}
\caption{CPT performance on \texttt{Llama2-7B}.}
\label{tab:cpt}
\end{table}

\begin{figure}[h!]
\centering
    \includegraphics[width=0.45\textwidth]{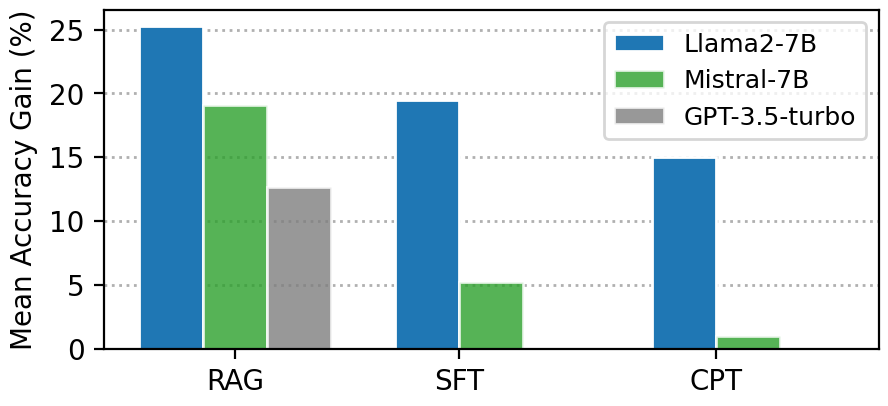}
    \vspace{-3mm}
    \caption{The relative performance (F1-score) gain for RAG, SFT, and CPT over averaged across all datasets.}
    \label{fig:rag_sft_cpt}
\end{figure}

\paragraph{Holistic Comparison of RAG, SFT, and CPT.} 
Fig.\ref{fig:rag_sft_cpt} presents a comprehensive comparison of the \skr method across various knowledge injection pipelines. While RAG demonstrates marked improvement over the raw model and baselines, the gains from CPT are relatively modest. Despite this, \texttt{Llama2-7B} consistently shows significant improvements over \texttt{Mistral-7B} across RAG, SFT, and CPT. \texttt{GPT-3.5-turbo} is utilized solely as the generator for RAG, where it exhibits relatively low-performance gains due to its already high base performance. Our \skr method appears promising in enhancing the RAG performance of open-source pre-trained models, suggesting that \texttt{Llama2-7B} could be an effective candidate for knowledge injection, potentially achieving competitive gains similar to RAG when applied through SFT and CPT.


\section{Related Work}


Incorporating new information into LLMs or refining their capabilities with previously seen information has posed a substantial challenge \cite{chen2022knowprompt,martino2023knowledge,zhang2023plug,ye2023qilin,zhang2023knowgpt}, the lack of effective approaches and evaluations is a notable concern. Recently, \citet{xu2023kilm} explored continued pre-training with a knowledge infill objective to enhance fact retention, while \citet{tian2023fine} applied constrained optimization through direct preference optimization \cite{rafailov2024direct} to reinforce factuality during post-training. Additionally, \citet{ovadia2023fine} aimed to compare retrieval, fine-tuning, and their combinations but the optimal way to inject knowledge into LLMs remains an open and sometimes contentious debate.

However, a critical knowledge ingestion gap remains: what are the data representations that work best for each injection pipeline with unprocessed and unstructured knowledge? \cite{baek2024knowledge} To answer this, \citet{balaguer2024rag} extracts information from raw PDF documents, leading to questions and answers for supervised fine-tuning pipelines. \citet{mecklenburg2024injecting} presents two data generation processes, token-based and fact-based, to inject knowledge via SFT. In contrast, our study aims to develop an innovative and generic synthetic knowledge ingestion approach, \skrnospace, which achieves significant improvement for each knowledge injection pipeline. 
\section{Conclusions}

We propose a novel method for refining knowledge representation and injection to enhance LLMs. Through extensive empirical analysis, our work has revealed several important findings. {\em Firstly}, synthetic QA generations are highly effective in digesting knowledge from its raw format, resulting in consistent enhancement for each injection pipeline. {\em Secondly}, fine-grained assemble augmentation has proven to be proficient in constructing high-quality, diverse datasets for knowledge injection. {\em Lastly}, RAG shows the potential to enhance knowledge, whereas the advancement of SFT and CPT largely depends on selecting appropriate base models and leveraging their existing knowledge capabilities.



\section{Ethics Statement}
This paper studies knowledge ingestion and representation in LLMs, which has significant broader impacts on the field of natural language processing and helps to address ethical considerations regarding factuality and reliability. The research outcome may contribute to the development of more accurate and factual LLMs by mitigating the risks of misinformation and biased outputs and promoting accountability and trust in AI systems.

\section{Limitations}
Our current experiments focus on tasks in a question-answering setting which requires a knowledge base. Further research is needed to expand to more complex tasks such as conversational or dialogue-based prompting. Also, the quality of the results introduced by \skr depends on the quality of the knowledge base. For OOD questions or edge cases where the knowledge base is corrupted or provides conflicting information, \skr is not able to solve that. \skr does require LLM calls to generate different representations. Such jobs can be conducted offline and only once. Thus the cost will be relatively reduced as it serves more queries. In addition, although advanced knowledge representations such as knowledge graphs show promise, their discussion falls beyond the scope of the current study but will appear in our future work.

\bibliography{custom}

\appendix
\onecolumn
\section{Appendix}
\label{sec:appendix}

\subsection{Synthetic Knowledge Pipeline Prompts}

\begin{table}[h]
\centering
\caption{Prompt to generate Questions for \skrqcnnospace}
\label{tab:qcn}
\begin{tabular}{p{13cm}}
\toprule
You are a question generator. Given a series of paragraphs, you have job is to generate questions for each paragraph.\\
\\
The question generated should be able to be answered ONLY based on the information in the paragraph. \\
The question generated should be about the main topic of the paragraph.\\
\\
\#\#Paragraphs:\\
\{\textit{paragraphs}\} \\
\\
Return the questions in a list. \\

["1. question 1", "2. question 2", "3. question 3" ...] \\
\\
\#\#Questions: \\
\bottomrule
\end{tabular}
\end{table}

\begin{table}[h]
\centering
\caption{Prompt to generate Questions And Answers for \skrqcannospace}
\label{tab:qcn}
\begin{tabular}{p{13cm}}
\toprule
You are a question generator. Given a series of paragraphs, you have job is to generate questions for each paragraph and abstract the corresponding answers.\\
\\
The question generated should be able to be answered ONLY based on the information in the paragraph. \\
The question generated should be about the main topic of the paragraph.\\
The answer should be about the generated question and based on the information in the paragraph.\\
\\
\#\#Paragraphs:\\
\{\textit{paragraphs}\} \\
\\
Return the questions in a list. \\

[\{"q": "question 1", "a": "answer 1"\}, \{"q": "question 2", "a": "answer 2"\}, ...] \\
\\
\#\#Questions: \\
\bottomrule
\end{tabular}
\end{table}

\clearpage

\subsection{Examples of Data generated under N-gram}
\label{sec:ngram}
\begin{longtable}[c]{ m{3cm} | m{12cm} }
\caption{Synthetic Data Example }
\label{tab:ngram}

 \endfirsthead

 \hline
 \multicolumn{2}{l}{Continuation of Table \ref{tab:ngram}}\\
 \hline

 \endhead
\hline
 \multicolumn{2}{l}{Continuation of Table \ref{tab:ngram}}\\
 \hline

 \endfoot

 \endlastfoot

 \toprule
\textbf{Configuration} & \textbf{Generated Synthetic Data} \\
 \midrule
 \textbf{Original Article} & Antonio Lucio Vivaldi (] ; 4 March 1678\u00a0\u2013 28 July 1741) was an Italian Baroque composer, virtuoso violinist, teacher and cleric. Born in Venice, he is recognized as one of the greatest Baroque composers, and his influence during his lifetime was widespread across Europe. He composed many instrumental concertos, for the violin and a variety of other instruments, as well as sacred choral works and more than forty operas.", "answer": "Antonio Vivaldi was a Baroque composer, virtuoso violinist, teacher and cleric.
\\ \midrule
 \textbf{\skrqcone} & [\{"\textbf{question}": " Who was Antonio Lucio Vivaldi?", \newline"\textbf{context}": "Antonio Lucio Vivaldi (] ; 4 March 1678\u00a0\u2013 28 July 1741) was an Italian Baroque composer, virtuoso violinist, teacher and cleric."\}, \newline \newline\{"\textbf{question}": " Where was Vivaldi born and what is he recognized for?", \newline "\textbf{context}": "Born in Venice, he is recognized as one of the greatest Baroque composers, and his influence during his lifetime was widespread across Europe."\}, \newline \newline \{"\textbf{question}": " What types of music did Vivaldi compose and what is his most famous work?", \newline"\textbf{context}": "He composed many instrumental concertos, for the violin and a variety of other instruments, as well as sacred choral works and more than forty operas."\}]
 \\ \midrule
  \textbf{\skrqctwo} &[\{"\textbf{question}": " Who was Antonio Lucio Vivaldi?", \newline"\textbf{context}": "Antonio Lucio Vivaldi (] ; 4 March 1678\u00a0\u2013 28 July 1741) was an Italian Baroque composer, virtuoso violinist, teacher and cleric. Born in Venice, he is recognized as one of the greatest Baroque composers, and his influence during his lifetime was widespread across Europe."\}, \newline \newline \{"\textbf{question}": " What types of music did Vivaldi compose?", \newline"\textbf{context}": "He composed many instrumental concertos, for the violin and a variety of other instruments, as well as sacred choral works and more than forty operas. His best-known work is a series of violin concertos known as \"The Four Seasons\"."\}]
   \\ \midrule
  \textbf{\skrqcthree} & [\{"\textbf{question}": " Who was Antonio Lucio Vivaldi?", \newline"\textbf{context}": "Antonio Lucio Vivaldi (] ; 4 March 1678\u00a0\u2013 28 July 1741) was an Italian Baroque composer, virtuoso violinist, teacher and cleric. Born in Venice, he is recognized as one of the greatest Baroque composers, and his influence during his lifetime was widespread across Europe. He composed many instrumental concertos, for the violin and a variety of other instruments, as well as sacred choral works and more than forty operas."\}, \newline \newline \{"\textbf{question}": " What type of music did Vivaldi compose?", \newline"\textbf{context}": "His best-known work is a series of violin concertos known as \"The Four Seasons\"."\}]
  \\ \midrule
\textbf{\skrqcaone} & [\{"\textbf{question}": "What was Antonio Lucio Vivaldi's profession?", \newline "\textbf{context}": "Antonio Lucio Vivaldi (] ; 4 March 1678\u00a0\u2013 28 July 1741) was an Italian Baroque composer, virtuoso violinist, teacher and cleric.", \newline"\textbf{answer}": "Antonio Lucio Vivaldi was a composer, virtuoso violinist, teacher and cleric."\},  \newline \newline \{"\textbf{question}": "Where was Antonio Lucio Vivaldi born?", \newline"\textbf{context}": "Born in Venice, he is recognized as one of the greatest Baroque composers, and his influence during his lifetime was widespread across Europe.", \newline "\textbf{answer}": "Antonio Lucio Vivaldi was born in Venice."\}, \newline \newline\{"\textbf{question}": "What is Antonio Lucio Vivaldi recognized as?", \newline"\textbf{context}": "He composed many instrumental concertos, for the violin and a variety of other instruments, as well as sacred choral works and more than forty operas.", \newline "\textbf{answer}": "Antonio Lucio Vivaldi is recognized as one of the greatest Baroque composers."\}, \newline \newline\{"\textbf{question}": "What types of works did Antonio Lucio Vivaldi compose?", \newline"\textbf{context}": "His best-known work is a series of violin concertos known as \"The Four Seasons\".", \newline "\textbf{answer}": "Antonio Lucio Vivaldi composed many instrumental concertos, sacred choral works, and more than forty operas."\}]
  \\ \midrule
\textbf{\skrqcatwo} & [\{"\textbf{question}": "What was Antonio Lucio Vivaldi's profession?", \newline "\textbf{context}": "Antonio Lucio Vivaldi (] ; 4 March 1678 - 28 July 1741) was an Italian Baroque composer, virtuoso violinist, teacher and cleric. Born in Venice, he is recognized as one of the greatest Baroque composers, and his influence during his lifetime was widespread across Europe.", \newline "\textbf{answer}": "Antonio Lucio Vivaldi was a composer, virtuoso violinist, teacher and cleric."\}, \newline \newline\{"\textbf{question}": "Where was Antonio Lucio Vivaldi born?", \newline "\textbf{context}": "He composed many instrumental concertos, for the violin and a variety of other instruments, as well as sacred choral works and more than forty operas. His best-known work is a series of violin concertos known as \"The Four Seasons\".", \newline"\textbf{answer}": "Antonio Lucio Vivaldi was born in Venice."\}]
  \\ \midrule
\textbf{\skrqcathree} & [\{"\textbf{question}": "What was Antonio Vivaldi's profession?", \newline "\textbf{context}": "Antonio Lucio Vivaldi (] ; 4 March 1678 - 28 July 1741) was an Italian Baroque composer, virtuoso violinist, teacher and cleric. Born in Venice, he is recognized as one of the greatest Baroque composers, and his influence during his lifetime was widespread across Europe. He composed many instrumental concertos, for the violin and a variety of other instruments, as well as sacred choral works and more than forty operas.", \newline "\textbf{answer}": "Antonio Vivaldi was a Baroque composer, virtuoso violinist, teacher and cleric."\}, \newline \newline \{"\textbf{question}": "Where was Antonio Vivaldi born?", \newline "\textbf{context}": "His best-known work is a series of violin concertos known as \"The Four Seasons\".", \newline "\textbf{answer}": "Antonio Vivaldi was born in Venice."\}]
\\ 
\bottomrule
 \end{longtable}


 \subsection{Details of SFT Pipeline}
 \label{sec:sft}
 \paragraph{Training.} For the SFT Pipeline, we use the QLoRA \cite{dettmers2024qlora} strategy with 4bit quantization built on top of Llama-Factory \cite{zheng2024llamafactory}\footnote{\url{https://github.com/hiyouga/LLaMA-Factory}}. For float format we use bf16. The batch size is set to 2, and the number of update steps to accumulate the gradients is set to 4. We use a cosine scheduler with a learning rate of 5e-5 and a warm-up ratio of 0.1. The training lasts for 3 rounds, with the maximum gradient norm set to 1. We use Amazon Sagemaker g5.12xlarge and g5.24xlarge instances for training, which are powered by NVIDIA A10 Tensor Core GPUs and comprise 4 GPUs with 24 GB of memory each. For training data size please refer to Table \ref{tab:training-data-size}.

 \paragraph{Generation.} We will generate 1 most likely generation with sampling strategies. This generation will be used to evaluate the correctness. The temperature of generation is fixed at 1.0, and top\_k is fixed at 50. The maximum number of new tokens in each generation is set to 40 tokens. Instruction used for generation is \textit{Respond to questions with concise and to-the-point answers. No explanation is needed. Keep your response within 20 words}. The test data for each dataset consists of 200 question-answer or question-context pairs.

 
\subsection{Details of CPT Pipeline}
 \label{sec:cpt}
\paragraph{Training.} For CPT, we use the QLoRA 4bit quantization with the LoRA+ strategy with a lambda value of 16.0 and float16 mixed precision training built on top of Llama-Factory \cite{zheng2024llamafactory}. The batch size is set to 2, and the number of update steps to accumulate the gradients is set to 4. We use a cosine scheduler with a learning rate of 5e-5 and a warmup ratio of 0.1. The training lasts for 3 rounds, with the maximum gradient norm set to 1. We use Amazon Sagemaker g5.12xlarge and g5.24xlarge instances for training, which are powered by NVIDIA A10 Tensor Core GPUs and comprise 4 GPUs with 24 GB of memory each.

\subsection{Details of Dataset} 

For each dataset, we select the first 200 queries in the test set as the final test set. Considering the huge amount of documents for retrieval, we select the articles that have been referred to at least one test query as the ground truth of the retrieval task to be the knowledge base, see Table \ref{tab:training-data-size}.

\begin{table}[h]
\caption{SFT/CPT Training Dataset Size.}
\centering
\begin{tabular}{l|cccc}
\toprule
\textbf{Dataset} & \textbf{\skrqaone} & \textbf{\skrqcone} & \textbf{\skrqaassemble} & \textbf{\skrqcassemble} \\ \hline
BioASQ \cite{tsatsaronis2015overview} & 5000 & 5000 & 34405 & 34966 \\ 
NQ \cite{kwiatkowski2019natural} & 5000 & 5000 & 13707 & 13528 \\ 
HotpotQA \cite{yang2018hotpotqa} & 5000 & 5000 & 12183 & 9191 \\ 
\bottomrule
\end{tabular}
\label{tab:training-data-size}
\end{table}

\begin{longtable}[c]{m{3cm}|m{12cm}}
\caption{CPT Training Data Example}
\label{tab:cpt-train-data}

 \endfirsthead
 \hline
 \multicolumn{2}{l}{Continuation of Table \ref{tab:cpt-train-data}}\\
 \hline
 \endhead
\hline
 \multicolumn{2}{l}{Continuation of Table \ref{tab:cpt-train-data}}\\
 \hline
 \endfoot
 \endlastfoot
 
\hline
\toprule
\textbf{Configuration} & \textbf{Generated Synthetic Data} \\
\midrule
\textbf{\skrqan}  & [\{"\textbf{text}": "Question: What was Antonio Lucio Vivaldi's profession?\newline \newline Answer: Antonio Lucio Vivaldi was a composer, virtuoso violinist, teacher and cleric."\},  \newline \newline \{"\textbf{text}": "Question: Where was Antonio Lucio Vivaldi born?\newline \newline Answer: Antonio Lucio Vivaldi was born in Venice."\}, \newline \newline\{"\textbf{text}": "Question: What is Antonio Lucio Vivaldi recognized as?\newline \newline Answer: Antonio Lucio Vivaldi is recognized as one of the greatest Baroque composers."\}] \\
\midrule
\textbf{\skrqcn} & [\{"\textbf{text}": "Question: Who was Antonio Lucio Vivaldi?"\newline \newline Context: "Antonio Lucio Vivaldi (] ; 4 March 1678\u00a0\u2013 28 July 1741) was an Italian Baroque composer, virtuoso violinist, teacher and cleric."\}, \newline \newline\{"\textbf{text}": "Question:  Where was Vivaldi born and what is he recognized for?\newline \newline Context: Born in Venice, he is recognized as one of the greatest Baroque composers, and his influence during his lifetime was widespread across Europe."\}, \newline \newline \{"\textbf{text}": "Question: What types of music did Vivaldi compose and what is his most famous work?\newline \newline Context: He composed many instrumental concertos, for the violin and a variety of other instruments, as well as sacred choral works and more than forty operas."\}] \\
\bottomrule
\end{longtable}

\paragraph{Generation.} We will generate 1 most likely generation with sampling strategies. This generation will be used to evaluate the correctness. The temperature of generation is fixed at 1.0, and top\_k is fixed at 50. The max number of new tokens of each generation is set to 40 tokens. Instruction used for generation is \textit{Respond to questions with concise and to-the-point answers. No explanation needed. Keep your response within 20 words}. The test data for each dataset consist of 200 question-answer or question-context pairs.

\begin{table}[h]
\centering
\caption{CPT Test Data Example}
\begin{tabular}{p{13cm}}
\toprule
\text{[}\{"\textbf{query}": "The 1978 NBA World Championship Series had as MVP which Hall of Fame class member of 1988?",\newline "\textbf{answer}": "Westley Sissel Unseld"\},\newline \newline \{"\textbf{query}": "Who was born first out of Leopold Lummerstorfer and Laurent Touil-Tartour?",\newline "\textbf{answer}": "Leopold Lummerstorfer"\}, \newline \newline \{"\textbf{query}": "In what city did Charlie Spiller play college football?",\newline "\textbf{answer}": "Lorman"\}\text{]} \\
\bottomrule
\end{tabular}
\end{table}

\clearpage

\subsection{Additional Experiments}
\subsubsection{Effect of N-gram on Retrieval} 
To gauge the impact of N-Gram on retrieval strategy, we compared the retrieval performance on FiQA task using the \skrqn strategy. Results can be observed in Table \ref{tab:n-gram-retreival}. 1-Gram strategy, despite the fact that it generates the most amount of questions to compare with, leads to the best retrieval result.

\subsubsection{Effect of Query Augmentation and Document Augmentation} 
To better understand the effectiveness of Query Augmentation and Document Augmentation on retrieval, we compare \skr with HyDE \cite{gao2022precise} using the FiQA dataset and Contriver. Full results can be seen in Table \ref{tab:hyde-retreival}. For the \skr variations, both \skrqn and \skrqcn outperform HyDE \cite{gao2022precise}.

\begin{table}[h!]
\centering
\caption{Effective of N-gram for Retrieval}
\label{tab:n-gram-retreival}
\begin{tabular}{l|ccc|ccc}
\toprule
        & \textbf{nCDG@1} & \textbf{nCDG@5} & \textbf{nCDG@10} & \textbf{Recall@1} & \textbf{Recall@5} & \textbf{Recall@10} \\
        \midrule
Article & 0.316 & 0.311 & 0.343  & 0.145 & \textbf{0.476}  & \textbf{0.562 }     \\
\midrule
3-Gram   & 0.451 & 0.428 & 0.456  & 0.225  & 0.328 & 0.429     \\
\midrule
2-Gram   & 0.448 & 0.435 & 0.468  & 0.221  & 0.454 & 0.554     \\
\midrule
1-Gram   & \textbf{0.452} & \textbf{0.454} & \textbf{0.482}  & \textbf{0.232} & \textbf{0.476}  & \textbf{0.562 }  \\  
\bottomrule
\end{tabular}
\end{table}

\begin{table}[h!]
\centering
\caption{Effective of Query Augmentation vs Document Augmentation}
\label{tab:hyde-retreival}
\begin{tabular}{l|ccc|ccc}
\toprule
        & \textbf{nCDG@1} & \textbf{nCDG@5} & \textbf{nCDG@10} & \textbf{Recall@1} & \textbf{Recall@5} & \textbf{Recall@10} \\
        \midrule
HyDE \cite{gao2022precise} & 0.427 & 0.442 & 0.478  & 0.221 & 0.472  & 0.576      \\
\midrule
\skrqone   & 0.452 & 0.454 & 0.482  & \textbf{0.232}  & 0.476 & 0.562     \\
\midrule
\skrqcone     & \textbf{0.455} & \textbf{0.458} & \textbf{0.493}  & 0.230 & \textbf{0.480}  & \textbf{0.584 }    \\
\midrule
\skrqcassemble   & 0.398 & 0.398 & 0.435  & 0.193 & 0.429  & 0.541   \\  
\bottomrule
\end{tabular}
\end{table}

\end{document}